\documentclass{article}

\usepackage[preprint]{corl_2023} % Uncomment for pre-prints (e.g., arxiv); This is like ``final'', but will remove the CORL footnote.

\usepackage{commenting}
\usepackage{subfigure}
\usepackage{amsmath}
\usepackage{amssymb}
\usepackage{graphicx}
\usepackage{wrapfig}
\usepackage[most]{tcolorbox}
\definecolor{cocoabrown}{rgb}{0.82, 0.41, 0.12}
\usepackage{tikz}
\newcommand{\tikzcircle}[2][color=red!40,fill=red!40]{\tikz[baseline=-0.5ex]\draw[#1,radius=#2] (0,0) circle ;}%
\usepackage{changepage}
\definecolor{ao(english)}{rgb}{0.0, 0.5, 0.0}
\definecolor{burntsienna}{rgb}{0.91, 0.45, 0.32}
\definecolor{bleudefrance}{rgb}{0, 0.5, 1}
\definecolor{etonblue}{rgb}{0, 0.8, 0}
\usepackage{varwidth}
\usepackage{tabularx}

\usepackage{booktabs}
\usepackage{tikz}

\title{Accelerating Reinforcement Learning of Robotic Manipulations via Feedback from Large Language Models}

\author{                                                             
  Kun Chu, Xufeng Zhao, Cornelius Weber, Mengdi Li, Stefan Wermter \\                                           
  Knowledge Technology, Department of Informatics, University of Hamburg, Germany\\
  \url{kun.chu@uni-hamburg.de} 
  }

\begin{document}
\maketitle

%===============================================================================

\begin{abstract}
Reinforcement Learning (RL) plays an important role in the robotic manipulation domain since it allows self-learning from trial-and-error interactions with the environment. Still, sample efficiency and reward specification seriously limit its potential. One possible solution involves learning from expert guidance. However, obtaining a human expert is impractical due to the high cost of supervising an RL agent, and developing an automatic supervisor is a challenging endeavor. Large Language Models (LLMs) demonstrate remarkable abilities to provide human-like feedback on user inputs in natural language. Nevertheless, they are not designed to directly control low-level robotic motions, as their pretraining is based on vast internet data rather than specific robotics data. In this paper, we introduce the \textbf{Lafite-RL} (\textbf{L}anguage \textbf{a}gent \textbf{f}eedback \textbf{i}n\textbf{te}ractive \textbf{R}einforcement \textbf{L}earning) framework, which enables RL agents to learn robotic tasks efficiently by taking advantage of LLMs' timely feedback. Our experiments conducted on RLBench tasks illustrate that, with simple prompt design in natural language, the Lafite-RL agent exhibits improved learning capabilities when guided by an LLM. It outperforms the baseline in terms of both learning efficiency and success rate, underscoring the efficacy of the rewards provided by an LLM.
\end{abstract}

% Submission to CoRL 2023 will be entirely electronic, via a web site (not email). Information about the submission process and \LaTeX{} templates are available on the conference web site at \url{https://corl2023.org/}. For camera ready submission, use the \texttt{final} option for the \texttt{\textbackslash usepackage} command. 

%===============================================================================

\section{Introduction}
Reinforcement Learning (RL) has shown its power in solving sequential decision-making problems in the robotic domain \citep{Arulkumaran2017, Kalashnikov2018}, through optimizing control policies directly from trial-and-error interactions with environments. However, there are still several challenges \citep{Ibarz2021}, like sample inefficiency and difficulties in specifying rewards, limiting its applications to the field. 

Inspired by how we human beings learn skills from more knowledgeable persons such as teachers or supervisors, a potential solution for the above limitations is learning from human expert guidance, so as to inject additional information into the learning process. Human guidance has shown some benefits in terms of providing additional rewards or guidance to accelerate the learning of new tasks, including learning from human demonstrations \citep{Ng2000, Arora2021} and feedback \citep{Knox2009, Wilson2012, Christiano2017, Warnell2018}. However, collecting sufficient human guidance is time-consuming and costly. 

Recently, Large Language Models (LLMs) have shown remarkable abilities to generate human-like responses in the textual domain \citep{Brown2020, Zhao2023}, and their applications have been explored in the robotic domain. While some approaches prompt LLMs to instruct robots in performing tasks \citep{Ahn2022, Vemprala2023, Ren2023, Zhao2023Matcha}, they focus on utilizing LLMs' common-sense knowledge to give high-level advice for employing pre-trained or hard-coded low-level control policies, which requires much data collection or expert knowledge respectively. Since these works do not perform policy learning when executing tasks with LLMs, the robots' performance highly depends on the LLM's capabilities and consistent presence during the interactions each time tasks are executed. Moreover, owing to their limited experience in the continuous 3D robotic world, LLMs are hard to employ for low-level robotic motion control. Direct learning low-level control policies taking advantage of feedback from LLMs is a promising solution, which leads to a research question: \textit{can LLMs understand scene information in robotic space and provide appropriate feedback to supervise reinforcement learning agents?}

\begin{figure*}[t]
    \subfigure[Human Feedback for RL]{
        \centering
        \includegraphics[width=0.49\linewidth]{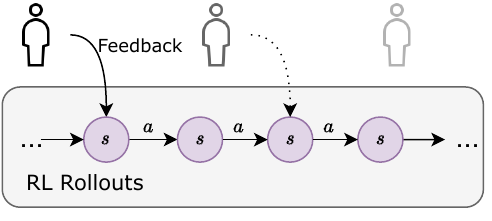}
    }
    \hfill
    \subfigure[Lafite-RL]{
        \centering
        \includegraphics[width=0.49\linewidth]{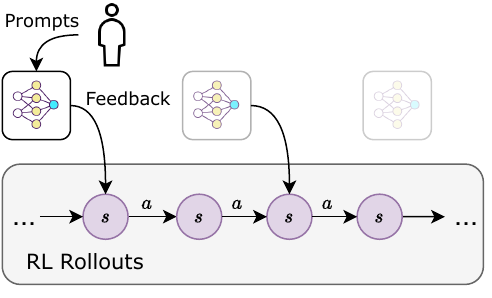}
    }
    \caption{Comparions of human feedback for RL and Lafite-RL. Normal human feedback for RL requires frequent human interaction during the learning process. In contrast, Lafite-RL allows a human to interact with the LLM only once, where the human prompts the LLM to observe the RL process and provide real-time feedback.}
    \label{fig:RLHF_vs_Lafite}
\end{figure*}

To this aim, we propose the \textbf{Lafite-RL} (\textbf{L}anguage \textbf{a}gent \textbf{f}eedback \textbf{i}n\textbf{te}ractive \textbf{R}einforcement \textbf{L}earning) framework that enables RL agents to efficiently learn user-specified robotic tasks with LLMs' real-time feedback, as shown in Fig.~\ref{fig:RLHF_vs_Lafite}. Before task learning, a human user with domain knowledge, but not necessarily robot programming knowledge, just needs to design two prompts: one to let the LLM understand the robotic scene, and the other to instruct it on how to evaluate the robot's behavior. Then, the LLM is able to provide interactive rewards mimicking human evaluative feedback, accelerating the RL process while freeing up human effort during the RL agent's numerous interactions with the task environment. Besides, the RL agent acquires information from the task environment and is robust to potential mistakes of the LLM. Experiments on RLBench tasks demonstrate our method's higher learning efficiency compared to the baseline method, which showcases that with easily designed prompts, an LLM's feedback is effective in accelerating RL in robotic manipulations.

\section{Related Works}

% \textcolor{red}{Waiting to be finished. More references will be added and analyzed.}

\subsection{Reinforcement Learning from Human Guidance} \label{RLHF_related_works}
There are several kinds of human guidance used for improving RL agents' ability to solve sequential decision-making tasks \citep{Zhang2021}, mainly including human demonstrations \citep{Arora2021} and feedback such as evaluative scores \citep{Knox2009} and preferences\citep{Christiano2017}.

\textbf{Learning from human demonstrations}. Inverse RL \citep{Ng2000, Ziebart2008, Arora2021} is an important technique of learning from human demonstrations, which focuses on inferring a reward function through observing collected human demonstrations. Then, the RL agent is able to learn policies efficiently with an estimated reward function that succinctly describes the current task. Other works combined imitation learning techniques with Deep RL \citep{Vecerik2017, Hester2018}, in which demonstrations are mostly used to initialize the agent's policy. However, those approaches normally require to collect high-quality demonstrations, which are not applicable to those task environments that are hard to demonstrate, e.g., when robots have a high degree of freedom. 

\textbf{Learning from human feedback}. Another set of helpful sources of information comes from diverse human feedback, mainly including evaluative scores \citep{Knox2009, Warnell2018, MacGlashan2017} and trajectory preferences \citep{Christiano2017, Lee2021, Wang2022}. Some works allow humans to observe state-action pairs when the agent interacts with the environment following the current policy, and provide feedback either explicitly like via keyboard keys \citep{Knox2009, MacGlashan2017} or implicitly like natural language \citep{Williams2018, Goyal2019}, gestures \citep{Yanik2013}, facial expressions \citep{Arakawa2018}, or their combinations \citep{Cruz2016, Trick2022}. Instead of evaluating state-action pairs, learning from human preferences allows human experts to give feedback on a set of trajectories in terms of rankings or preferences \citep{Christiano2017, Lee2021}. Those forms of feedback are interactively collected and then injected into the RL processes in terms of rewards \citep{Knox2009, Christiano2017, Goyal2019} or policies \citep{Griffith2013, MacGlashan2017, Arumugam2019}. While the existence of numerous forms of human feedback, the quality of human feedback is also an important aspect when discussing how to inject external knowledge into the RL process \citep{Griffith2013, Cruz2018}. Recently, learning from human preferences has also shown its power in fine-tuning LLMs to make them generate more human-like responses \citep{Ouyang2022}. Despite the successes shown in the works mentioned above, they require a significant amount of human feedback, which is normally time-consuming and expensive to collect.

\subsection{Large Language Models in Robotics}
The recent advancements in LLMs exhibit their remarkable abilities to generate complex text on a wide range of topics \citep{Zhao2023, Wei2022}. Likewise, some works utilize LLMs' zero-shot reasoning capabilities to guide agents on robotic tasks \citep{Ahn2022, Vemprala2023, Ren2023, Zhao2023Matcha, Liang2023, Yu2023}. With several skills libraries either manually designed by experts or pertained on massive data, these approaches do not perform any model training but instead use LLMs as planners for giving advice in a high-level manner. In this sense, robots’ performance highly depends on the LLM’s capabilities and requires the LLM's presence each time when the task is performed. Meanwhile, due to limited text corpora in the continuous 3D robotic world, LLMs struggle with direct low-level robotic motion control. In contrast to theirs, our framework aims to prompt the LLM to provide feedback to train RL-based models for use-specified tasks. After the training, the RL-based model is able to perform the task with a higher success rate without LLM's feedback or response. 

Several works use LLMs to provide rewards for RL. \citet{Yu2023} propose to prompt LLMs to define reward functions with specific parameters to optimize control policies on a variety of robotic tasks. However, their method relies on much manual work by experts, and the reward function is domain-specific and thus can not be applicable to broader domains. In contrast, our framework enables a normal human user to easily design the prompt, and interactive rewards are easy to understand and apply to various task domains. While \citet{Lee2023} and \citet{Kwon2023} demonstrate the effectiveness of employing LLMs' feedback in terms of preferences and evaluative scores in the RL context, they are mainly on textual world tasks like text summarizations \citep{Lee2023} and negotiation games \citep{Kwon2023}. In contrast, we aim to enable RL agents to benefit from LLMs' feedback when training on robotic manipulations, which requires LLMs' reasoning ability to understand the robot's behavior in continuous 3D space.

\section{Language Agent Feedback Interactive Reinforcement Learning}
\begin{figure*}[t]
\centering
\includegraphics[width=.85\linewidth]{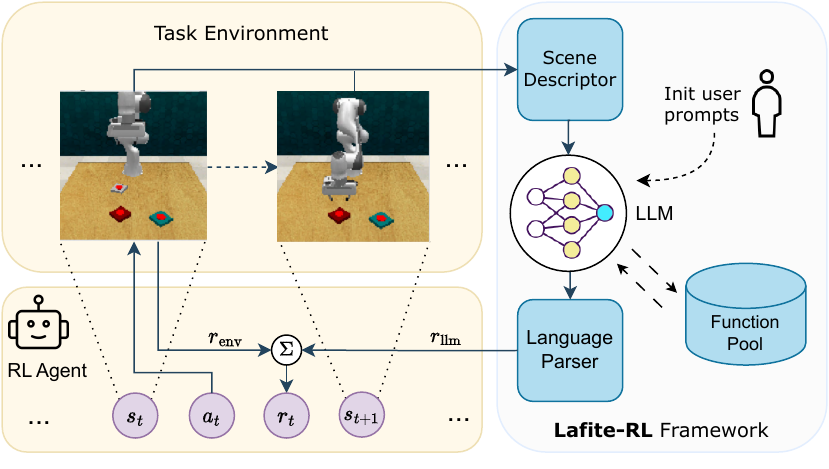}
\caption{Depiction of proposed Lafite-RL framework. Before learning a task, a user provides designed prompts (see Table \ref{tab:prompts}), including descriptions of the current task background and desired robot's behaviors, and specifications for the LLM's missions with several rules respectively. Then, \textbf{Lafite-RL} enables an LLM to ``observe" and understand the scene information which includes the robot's past action, and evaluate the action under the current task requirements. The language parser transforms the LLM response into evaluative feedback for constructing interactive rewards. } 
\label{fig:LafiteRL}
\end{figure*}
We formulate the learning agent's interaction with the task environment in RL as a Markov Decision Process (MDP) in terms of a tuple $(\mathcal{S}, \mathcal{A}, \gamma, \mathcal{T}, \mathcal{R})$, where $\mathcal{S}$ is the set of states, $\mathcal{A}$ is the set of actions, $\gamma \in [0,1]$ is the discount factor, $\mathcal{T}: \mathcal{S} \times \mathcal{A} \times \mathcal{S} \rightarrow [0,1]$ is the transition probability function, and $\mathcal{R}$ is the reward function that maps the state-action-state pair to a numerical reward $r: \mathcal{S} \times \mathcal{A} \times \mathcal{S} \rightarrow \mathbb{R}$, such that at timestep $t$, the RL agent receives a reward $r_{t}$ when transiting to state $s_{t+1}$ after executing an action $a_{t}$ at state $s_{t}$. By interacting with the environment, an RL algorithm seeks to learn an optimal policy $\pi$ that maximizes the accumulated rewards.

% specifying the probability $p(s^{\prime}|s, a)$ of reaching state $s' \in \mathcal{S}$ after taking action $a$ in state $s$,

Inspired by the practice of letting humans provide evaluative feedback for the agent's behaviour as in Section \ref{RLHF_related_works}, we propose the \textbf{Lafite-RL} framework that enables an LLM to generate real-time interactive rewards like a ``human teacher" to accelerate the robot's RL process. As shown in Fig.~\ref{fig:LafiteRL}, with designed prompts, for the request at timestep $t$, our framework enables the LLM to ``observe'' the transition $s_{t} \rightarrow s_{t+1}$ and understand the robot's motion $a_{t}$ under the current task requirement and generate an evaluation for it. Then, with a language parser, the textual response is transformed to a numerical value and injected into the RL process.

\textbf{LLM as Scene Observer}. We develop a scene descriptor to describe the scene based on the original information from the environment, including coordinates of the gripper and objects, the gripper's opening/closing state, and the gripper sensor information (if some object is detected or grasped). This description illustrates the movement from the previous state to the current state. We provide a prompt to the LLM that contains a function API: \textit{distance(objectA, objectB)} since the LLM requires geometric information from the environment. In this way, the LLM is required to understand the robot's motion under the current task requirement, and then make several function calls that can help itself for further evaluation. Based on the original information from the environment, and some outputs after the function calls, the LLM is prompted to give the final evaluation of the robot's motion. 

\textbf{LLM as Motion Evaluator.} For the evaluation, LLM is prompted to infer which phase the agent is during this period, and then evaluate if the current motion is good or not. A long-horizon manipulation task consists of many steps that belong to several phases, i.e., the gripper is required to approach the desired object, grasp the desired object, and lift the desired object to the given place. Therefore, the LLM needs to estimate which phase it is currently in since the optimal behaviors for each phase are different. Then, the LLM will generate its judgment for this motion in terms of \textit{good move} or \textit{bad move}, which is formed into a numerical score by the language parser. Formally, the reward for the RL agent at time $t$ is a combination of environment reward $r_{t}$ and scaffolding reward $r_{llm}$ by an LLM:
% \begin{equation}
% r_{t} = \left\{ 
% \begin{array}{ll}
% r_{\text{env}} + r_{\text{llm}} & \mbox{if query LLM} \\
% r_{\text{env}} & \mbox{else} \\
% \end{array}\right.
% \end{equation}
\begin{equation}
r_{t} = r_{\text{env}} + r_{\text{llm}.} 
\end{equation}
% \xf{BEGIN DELETE}
% where $r_{\text{env}}$ is the reward obtained from the environment, and $r_{\text{llm}}$ is the reward from the LLM provided via the language parser as:
% \xf{END DELETE}
% \xf{
The additional reward $r_{\text{llm}}$ is parsed from the LLM response, mapped as
% }
\begin{equation}
r_{\text{llm}} = \left\{ 
\begin{array}{ll}
1, & \mbox{if \textit{good move},} \\
-1, & \mbox{if \textit{bad move},} \\
0, & \mbox{else.}
\end{array}\right.
\end{equation}
The language parser interprets the response of the LLM based on keyword detection. In case it fails to identify the LLM output as \textit{good move} or \textit{bad move}, it will pass on zero to the agent.

Lafite-RL provides an interface for human users' prompt engineering and RL agents' training with LLMs' continual real-time feedback. Table \ref{tab:prompts} shows snippets of prompts designed in Lafite-RL. Instead of explicitly designing the reward function, the human user describes the requirements of the task in natural language and guides the LLM to understand the scene information and evaluate the robot's motion accordingly.
\begin{table}[htp]
    \centering
    \caption{Snippets of the two prompts required by Lafite-RL for learning one task. The missions of Scene Observer (\tikzcircle[color=etonblue!55, fill=etonblue!80]{2.5pt}) and Motion Evaluator (\tikzcircle[color=bleudefrance!55, fill=bleudefrance!55]{2.5pt}) are highlighted with different colors in the prompts.}
    \begin{tabular}{c}
    \includegraphics[width=0.98\linewidth]{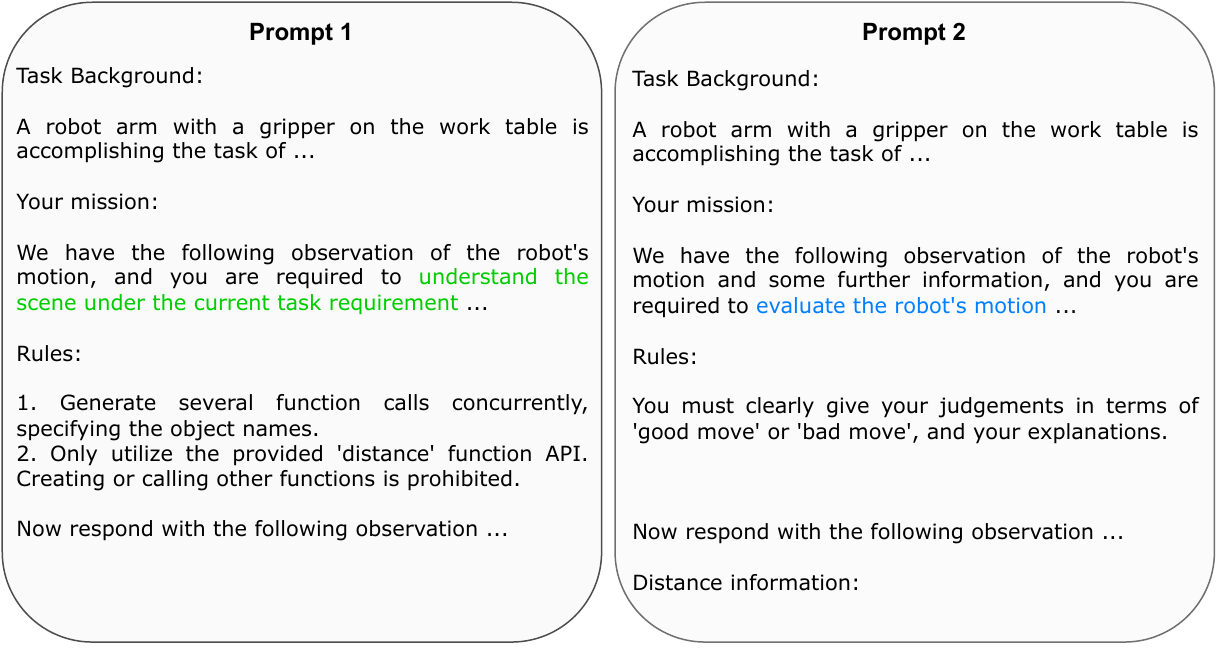}
    \end{tabular}
    \label{tab:prompts}
\end{table}
\section{Experiments}
We evaluate \textbf{Lafite-RL} to determine the extent to which an LLM can assist RL for robotic tasks in terms of evaluative feedback. For the LLM setup, we choose Vicuna-13B v1.5 \citep{Vicuna2023} as the underlying LLM model, which is a lightweight open-source model fine-tuned on the LLaMA-2 \citep{Touvron2023} model. For higher throughput and faster response time, we adopt the technique of vLLM \citep{Kwon2023vLLM} as an augmentation of our local LLM. For the task learning setup, we chose a state-of-the-art RL algorithm, DreamerV3 \cite{Hafner2023} as the underlying RL algorithm, and conducted experiments on RLBench \cite{James2020}, a suite of large-scale benchmark and learning environments designed for robot manipulation tasks. Tasks in RLBench are viewed as challenging for RL agents, as their rewards are sparse, i.e., the agent only obtains scores on the goal state and zero otherwise. 
\begin{figure*}[b]
\centering
    \subfigure[]{
        \centering
        \includegraphics[width=0.33\linewidth]{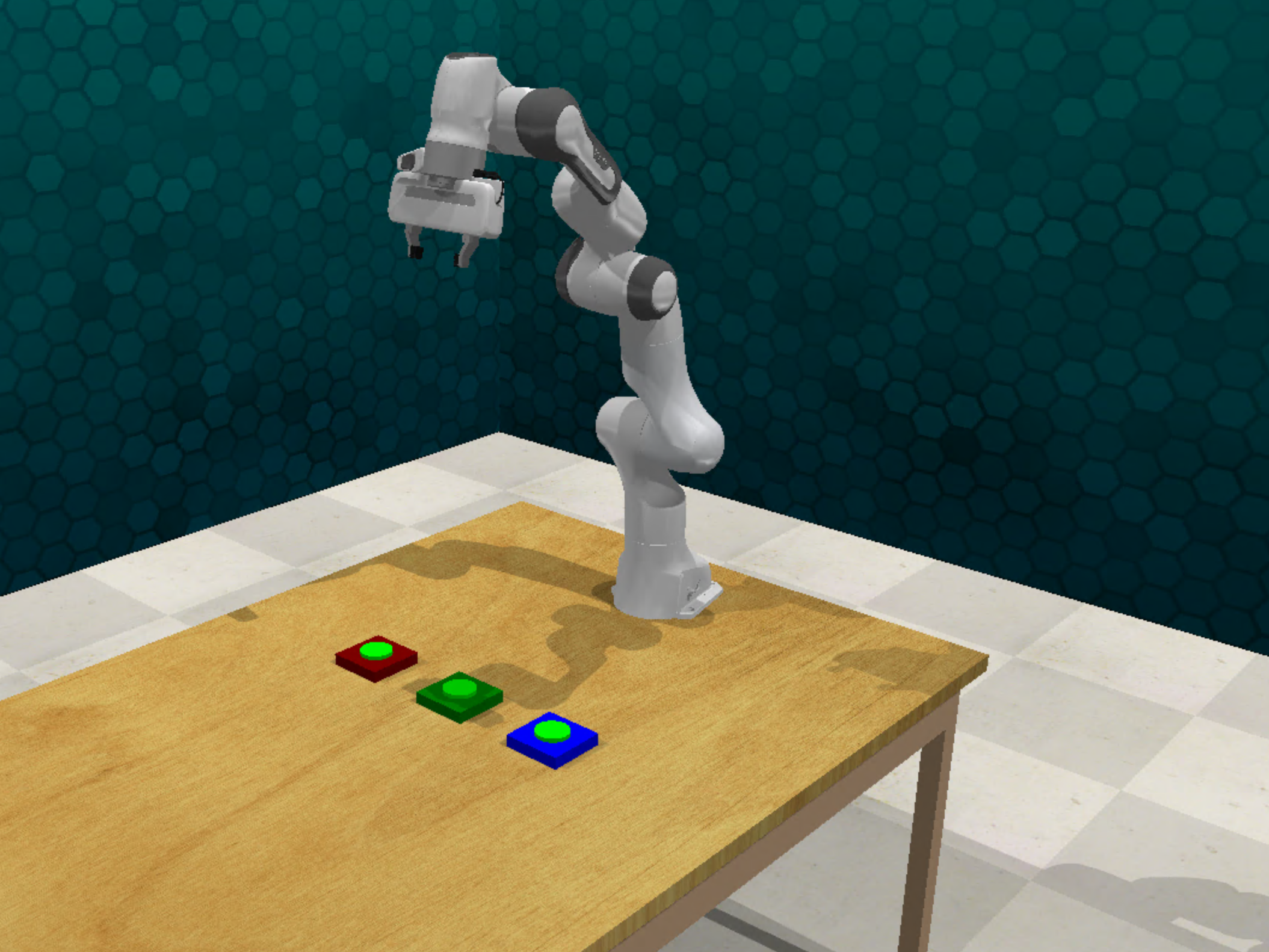}
    }\:\:\:\:\:\:
    \subfigure[]{
        \centering
        \includegraphics[width=0.33\linewidth]{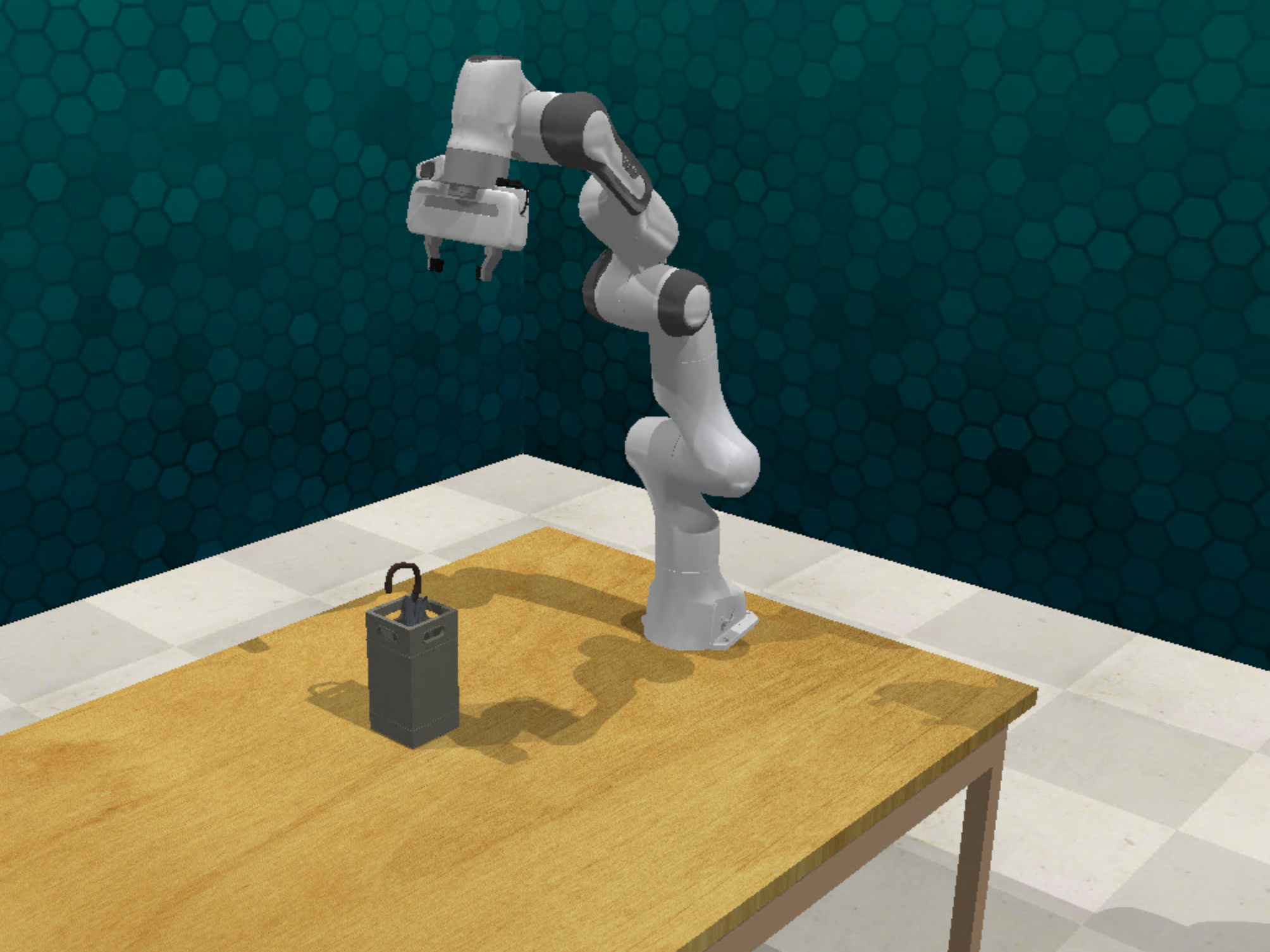}
    }%
\caption{Illustration of two RLBench tasks for evaluation: (a) Push buttons, and (b) Take umbrella out of umbrella stand.}
\label{fig:RLBench_Tasks}
\end{figure*}
Our environmental setup consisted of a Franka Emika Panda robot with one camera each on the wrist, front, and overhead sides, respectively. The action space is 4-dimensional, including delta values in x, y, and z directions as well as the manipulation of the gripper, and its observation space combines the visual space of the three camera inputs and the linear space of the gripper's low-dimensional state. 

% \begin{figure*}[htp]
%     \subfigure[Success rate during training]{
%         \begin{minipage}[t]{0.49\linewidth}
%             \centering
%             \includegraphics[width=1\linewidth]{figures/Success_rates_push_buttons.png}
%         \end{minipage}%
%         \begin{minipage}[t]{0.49\linewidth}
%             \centering
%             \includegraphics[width=1\linewidth]{figures/Success_rates_umbrella.png}
%         \end{minipage}%
%     }

%     \subfigure[Rewards obtained during training]{
%         \begin{minipage}[t]{0.49\linewidth}
%             \centering
%             \includegraphics[width=1\linewidth]{figures/Rewards_push_buttons.png}
%         \end{minipage}%
%         \begin{minipage}[t]{0.49\linewidth}
%             \centering
%             \includegraphics[width=1\linewidth]{figures/Rewards_umbrella.png}
%         \end{minipage}%
%     }

%     \subfigure[Episode length during training]{
%         \begin{minipage}[t]{0.49\linewidth}
%             \centering
%             \includegraphics[width=1\linewidth]{figures/Episode_length_push_buttons.png}
%         \end{minipage}%
%         \begin{minipage}[t]{0.49\linewidth}
%             \centering
%             \includegraphics[width=1\linewidth]{figures/Episode_length_umbrella.png}
%         \end{minipage}%
%     }
    
%     \caption{Learning curves of Lafite-RL (red curves) and the baseline without interactive feedback (grey curves) during training on two RLBench tasks. (a) success rates, and (b) rewards obtained during training.}
%     \label{fig:Learning_surves}
% \end{figure*}
\begin{figure*}[t]
    \subfigure[Push buttons]{
        \begin{minipage}[t]{0.33\linewidth}
            \centering
            \includegraphics[width=1\linewidth]{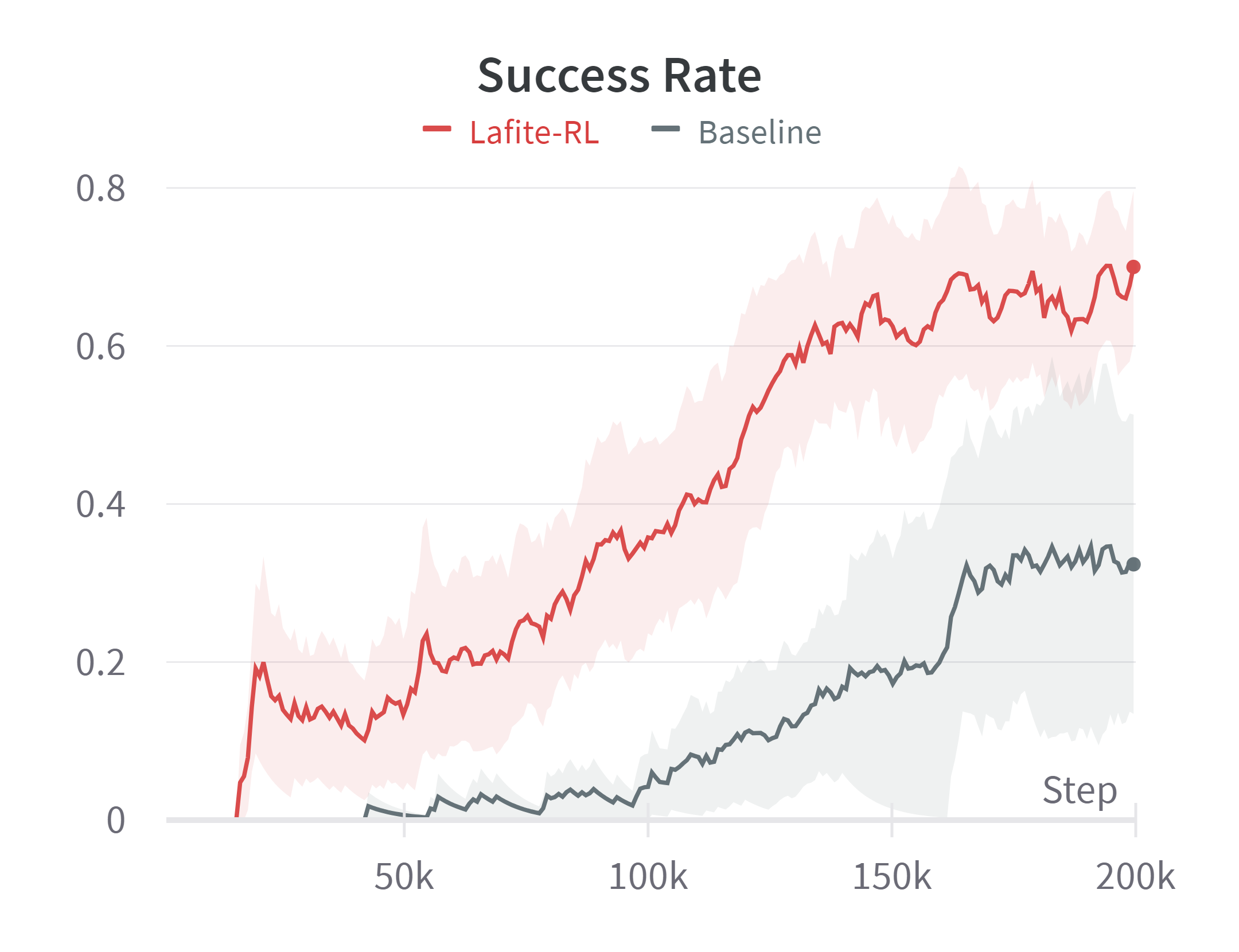}
        \end{minipage}
        % \hfill
        \begin{minipage}[t]{0.33\linewidth}
            \centering
            \includegraphics[width=1\linewidth]{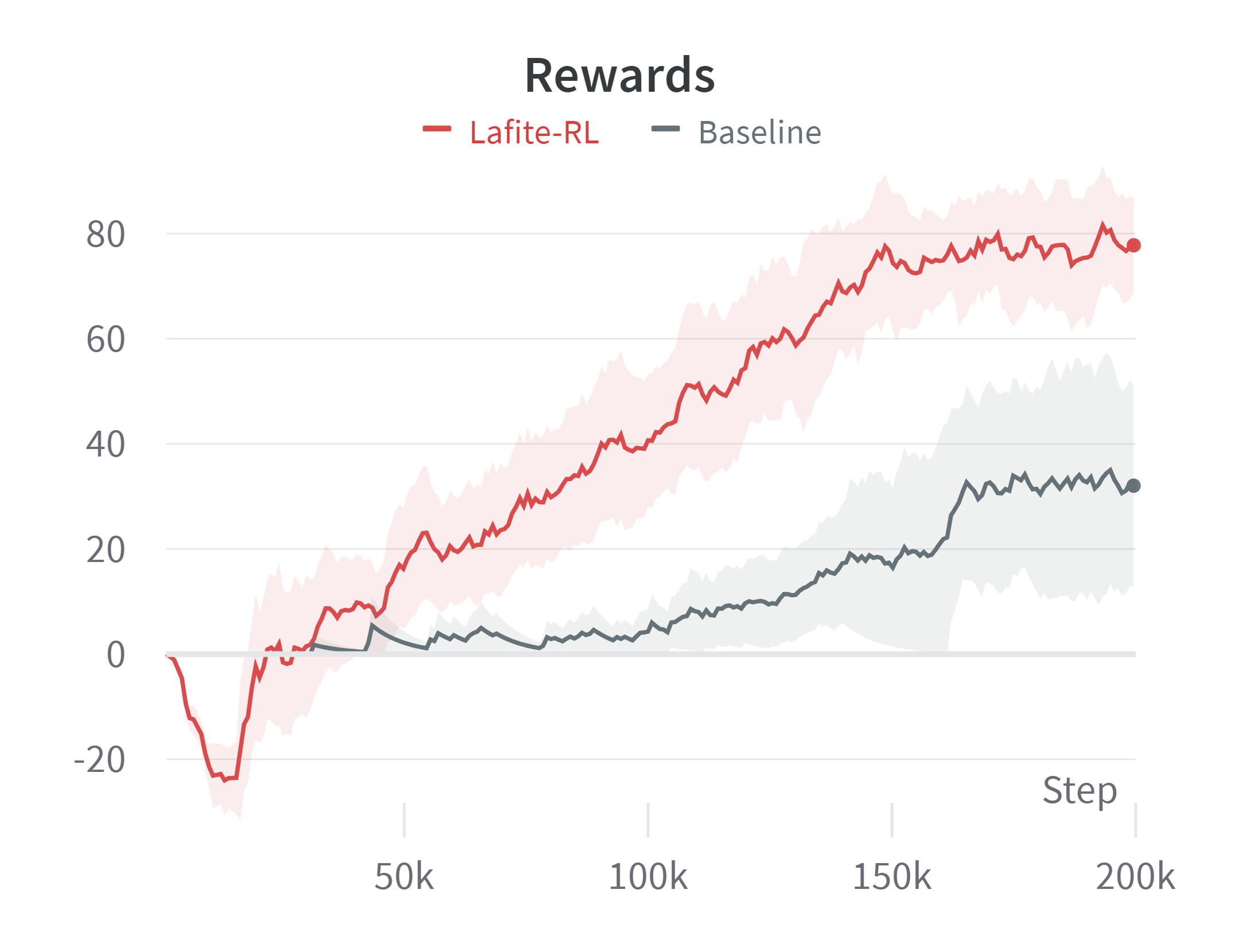}
        \end{minipage}
        % \hfill
        \begin{minipage}[t]{0.33\linewidth}
            \centering
            \includegraphics[width=1\linewidth]{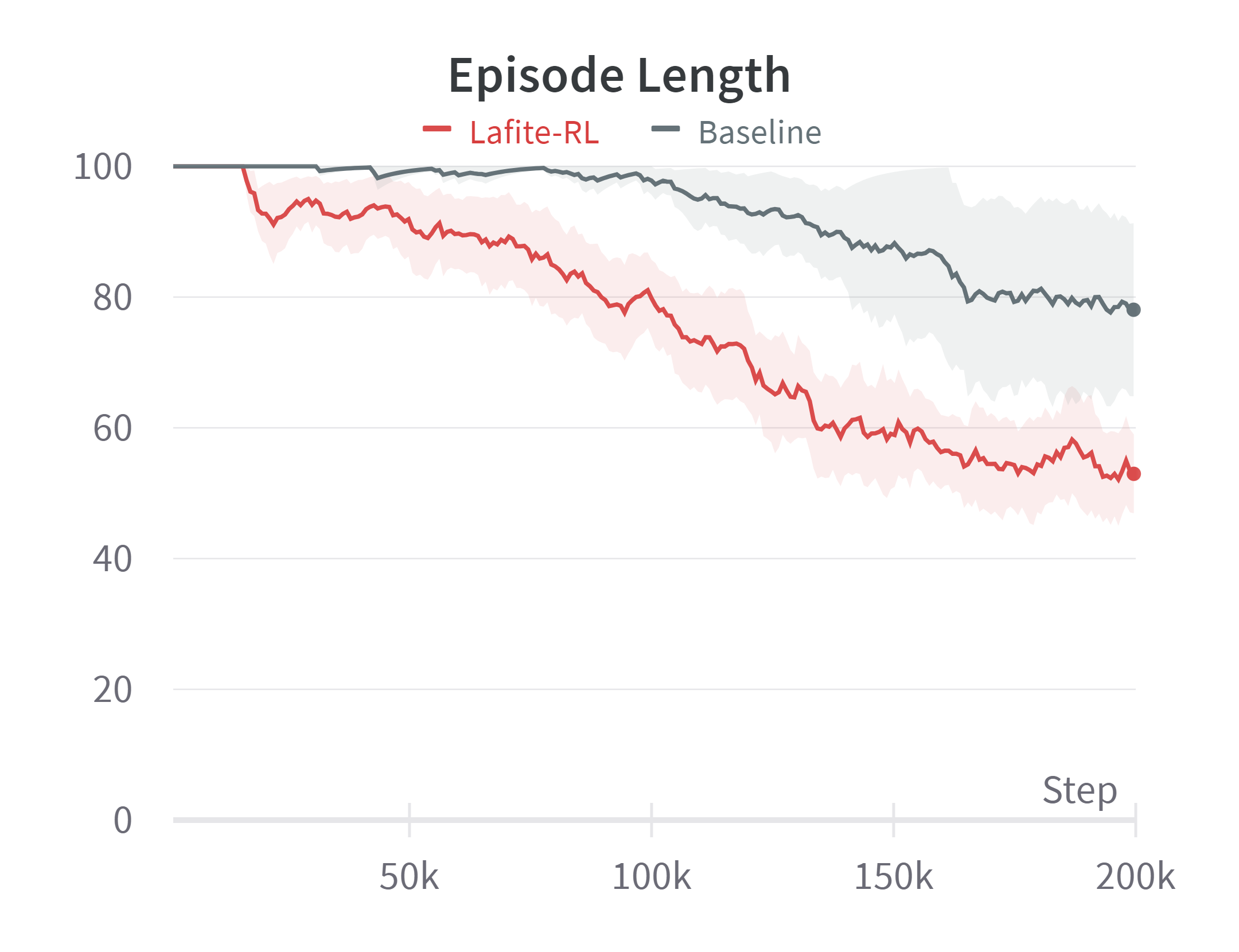}
        \end{minipage}
    }

    \subfigure[Take umbrella out of umbrella stand]{
        \begin{minipage}[t]{0.33\linewidth}
            \centering
            \includegraphics[width=1\linewidth]{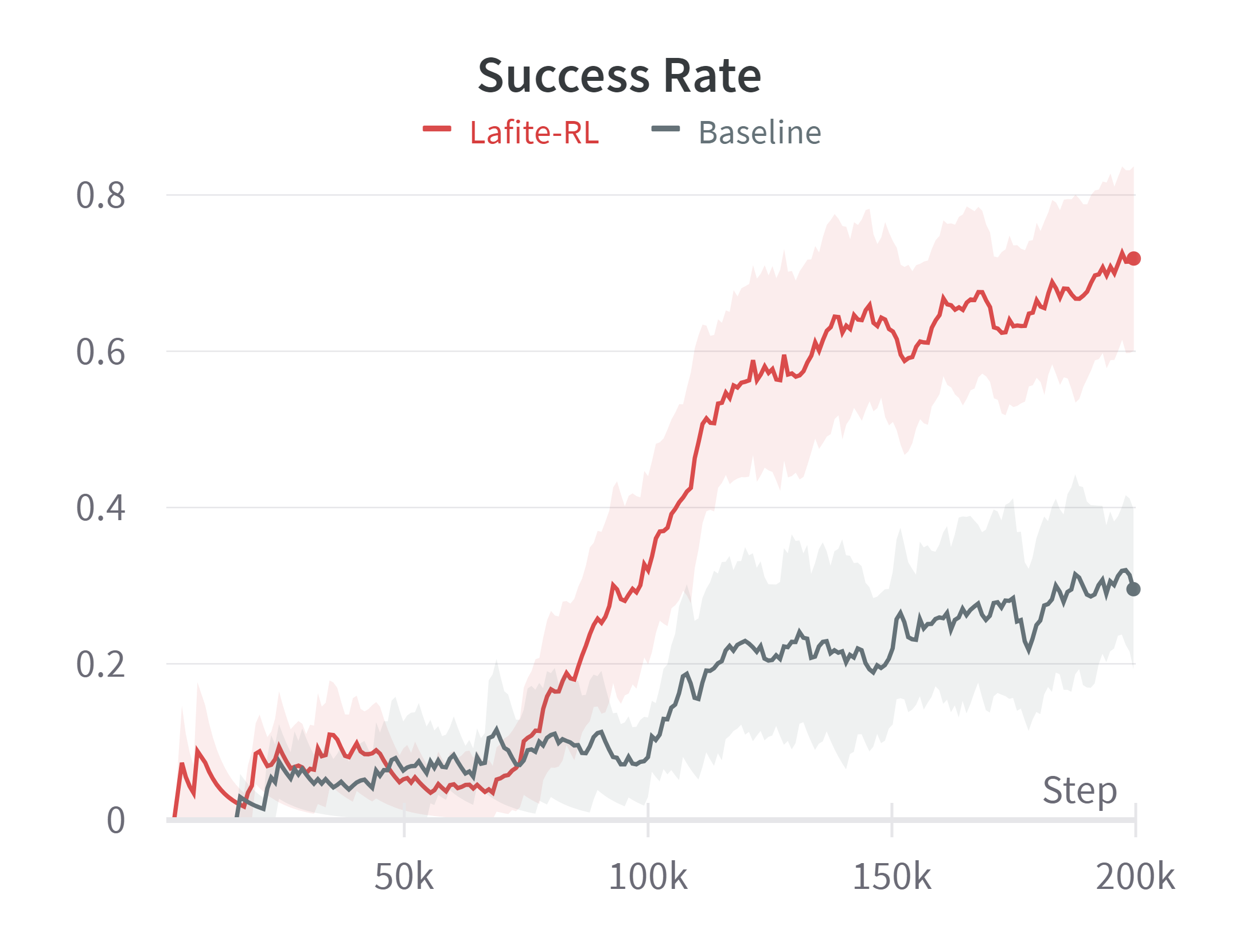}
        \end{minipage}
        \hfill
        \begin{minipage}[t]{0.33\linewidth}
            \centering
            \includegraphics[width=1\linewidth]{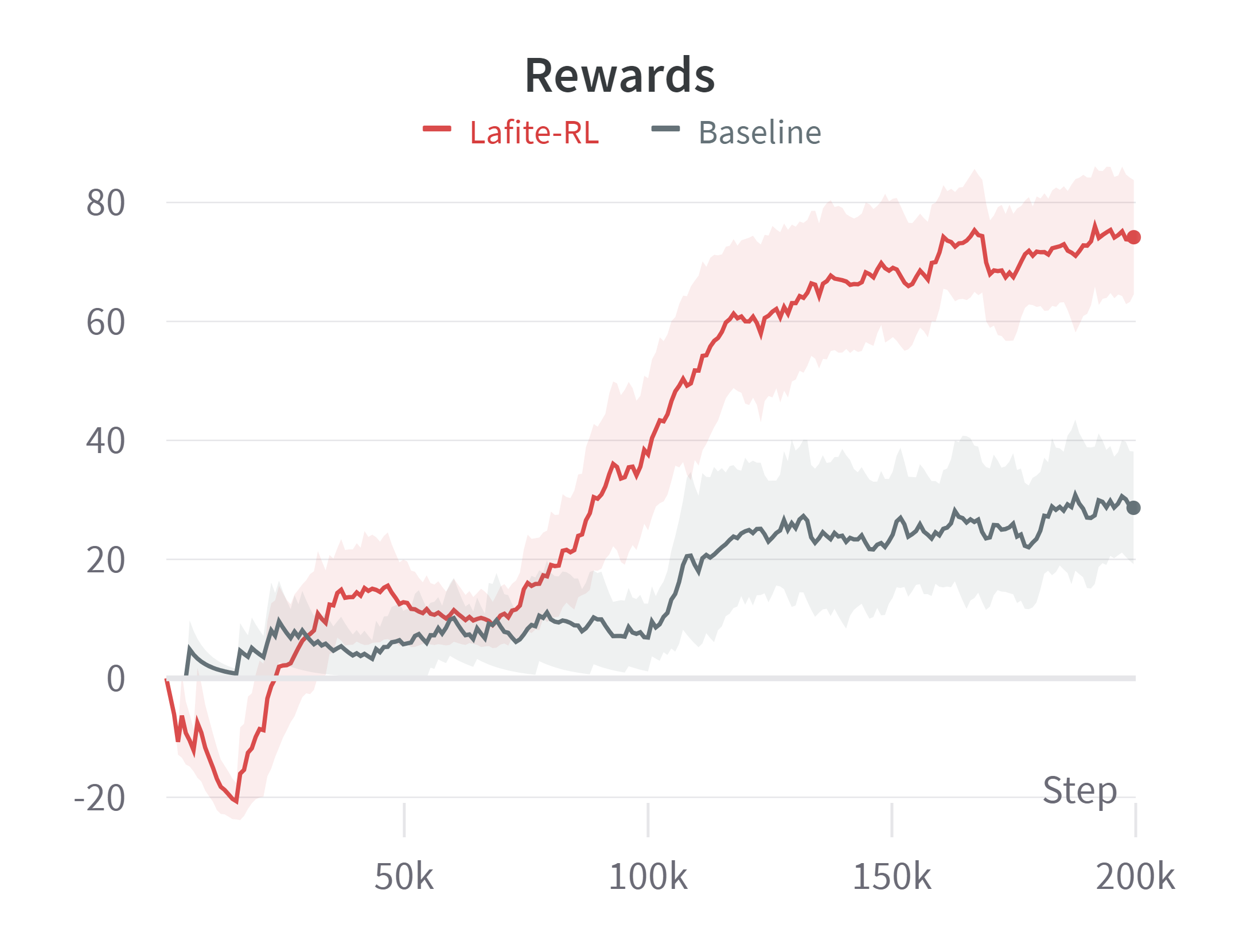}
        \end{minipage}
        \hfill
        \begin{minipage}[t]{0.33\linewidth}
            \centering
            \includegraphics[width=1\linewidth]{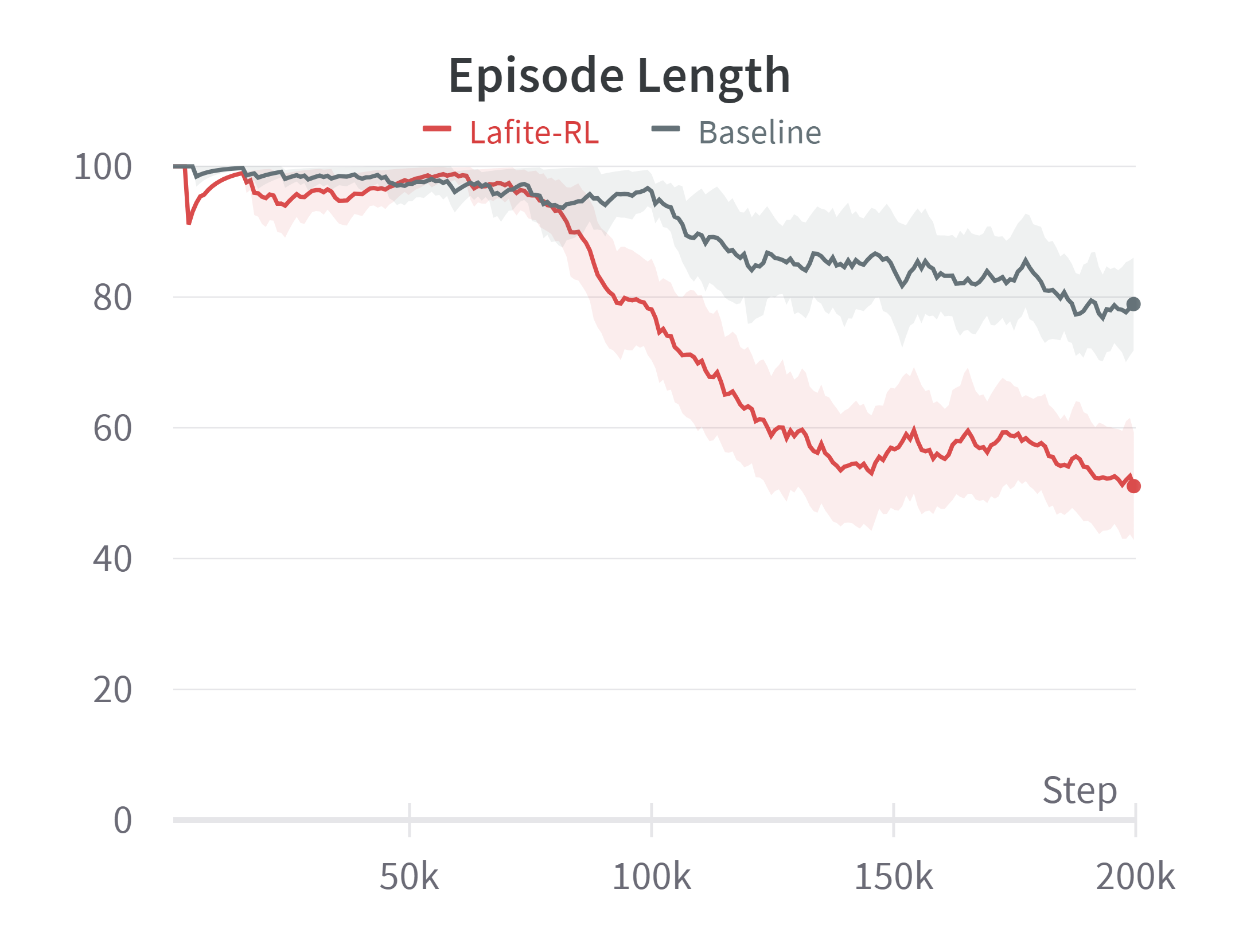}
        \end{minipage}
    }
    \caption{Learning curves of Lafite-RL (red curves) and the baseline without interactive feedback (gray curves) during training on two RLBench tasks. }
    \label{fig:Learning_surves}
\end{figure*}

We choose two RLBench tasks in experiments, depicted in Fig. \ref{fig:RLBench_Tasks}, \textit{Push buttons} and \textit{Take umbrella out of umbrella stand}. Positions of buttons or umbrellas are randomly initialized for each episode. In all experiments, we set the maximum episode length as 100, the maximum training steps as 2e5, the reward scale for the task's goal state to 100, and the frequency of querying the LLM as every 4 timesteps. We choose the middle-size version of Dreamerv3 and keep other hyperparameters as defaults in \citep{Hafner2023}. Each experiment was repeated three times with different random seeds. 

\begin{table}[ht!]
\centering
\caption{Success rates when evaluating different methods on two RLBench tasks for 100 episodes.}
% \begin{small}
\footnotesize \setlength{\tabcolsep}{3pt}
\begin{tabularx}{0.54\linewidth}{lcc}
\toprule
\textbf{Task Success Rates} & Lafite-RL & Baseline \\
\midrule
Push buttons & 70.3\% & 41.6\% \\
Take umbrella out of umbrella stand & 68.7\% & 26.1\% \\
\bottomrule
\end{tabularx}
\label{tab:success_rate}
\end{table}
In order to examine the influence of the interaction of the LLM, we report the learning curves regarding rewards, success rates, and episode length during the training process in Fig.~\ref{fig:Learning_surves}. Tab.~\ref{tab:success_rate} shows the success rate of each RL policy over 100 evaluation episodes after learning. 

We observe that, for the \textit{Push buttons} task, success rates performed by Lafite-RL witness a jump at 20k steps and then continue to rise, while the rewards drop at the beginning, nadir at around 12k steps, and then continue to rise. In the case of the ``take umbrella" task, the RL agent initially faces challenges in its learning process, mainly due to the complexity of the task. Later Lafite-RL agent has a substantial increase compared to the baseline, indicating that, especially for a long-horizon task that requires a larger number of steps to accomplish (cf. Fig.~\ref{fig:Learning_surves} right column), the guidance of an LLM proves to be particularly advantageous.
In contrast, the Baseline did not perform well in two tasks in terms of slow-rising speed and low success rates and rewards. By the end of the training, Lafite-RL is able to achieve episode to be an average of 50 in both tasks, while the Baseline can only get around 80, which demonstrates the effectiveness of the control policy learned by Lafite-RL. During the evaluations after training, as shown in Table \ref{tab:success_rate}, Lafite-RL shows a much higher success rate on 100 episodes of evaluations, which showcases that after training with LLMs' continual real-time feedback, RL agents are able to perform the task well independently. 

\section{Discussion}

Our experiments have shown that an LLM can successfully provide interactive feedback, similarly as a human can do it, to significantly enhance RL. There are other strategies to accelerate RL, however, each incurs costs or relies on preconditions to be satisfied, for example, RLHF requires a human to be continually present; imitation learning requires trajectories being performed veridically beforehand; reward shaping requires suitable knowledge of the solution strategy. When these techniques are unavailable, or as a complementary technique, Lafite-RL offers a new alternative. However, it incurs its own costs, such as the need to prompt an LLM and frequent time-consuming interactions between the agent and the LLM. In our implementation, the agent requests LLM feedback at fixed-time steps to trade off costly feedback vs.\ cheaper feedback-less phases. More sophisticated strategies could let the agent dynamically request feedback only when it is expected to be particularly useful \citep{torrey2013teaching}. At the same time, to use the LLM more efficiently, in addition to evaluative feedback as in this study, more informative feedback from LLMs like action instructions will be needed to guide the RL process. 

In the current Lefite-RL framework, a motion description is made based on the coordinates of objects in the environment, which might not be accessible in some environments, and other information like the gripper's angles or velocities is not involved.
More spatial-related tools can be designed to serve as APIs to let LLMs better understand the scene in complex task environments.
Currently, several vision-language foundation models are being developed \citep{Radford2021, gan2022vision}. Their use can augment the LLM's understanding of the environment with more useful information, so as to provide better judgment and instructions.
These visual capabilities will aid in the transfer from simulation to the real world.

From an experimental perspective, it is desirable to conduct more experiments with a larger variety of tasks, as well as evaluate the model with respect to generalization to model-free RL algorithms. Towards deploying the model, studies with non-expert users as well as sim-to-real or training in the real world would also be further steps.

\section{Conclusion}
In this work, we introduce Lafite-RL, a framework aimed at enabling non-expert users to design prompts that instruct LLMs to give timely feedback during RL agents' training in robotic tasks, so as to accelerate learning while freeing up human effort. Experimental results show that with easily designed prompts, an LLM's feedback is effective in speeding up reinforcement learning interactively.
Since LLMs are known to be skilled in high-level actions, our results in the low-level robotic manipulation domain demonstrate the potential of LLMs, where we have not used them directly for control but to guide an agent's learning. In summary, our approach shows substantial potential for utilizing LLMs for interactively enhancing reinforcement learning. 

% future work: extension to instructive feedback; more experiments with further tasks, generalization to model-free algorithnms; user studies needed with real non-expert users in real scenes, sim-to-real; benefit from vision-language models, ...
% Meanwhile, as the performance of internet-scale vision-language models \citep{Radford2021} improves, we expect that such foundation models will continually add more vibrancy to the RL community.   will let the LLM understand the scene more easily, making our model more easily deployable.

% which reveals two interesting challenges, scene and motion description, and forms of feedback. In the future, in addition to evaluative feedback, we will extend our model to using instructive or corrective feedback.

% diverse forms of feedback from LLMs will be used for guiding the RL process in a broader context of robotic tasks with accurate motion descriptors. 

%===============================================================================

% \clearpage
% The acknowledgments are automatically included only in the final and preprint versions of the paper.
\acknowledgments{We gratefully acknowledge support from the China Scholarship Council (CSC) and the German Research Foundation (DFG) under the project Crossmodal Learning (TRR 169).
}

%===============================================================================
\clearpage
\bibliography{references}  % .bib

\end{document}